# Neural Net Model for Featured Word Extraction

**A. Das**
Department of Mathematics, Jadavpur University,
Calcutta 700 032India; email: atin_das@yahoo.com
**M. Marko**
Faculty of Management, Comenius University, Slovakia
**A. Probst**
Faculty of Mathematics, Physics & Informatics,
Comenius University, Slovakia
**M. A. Porter**
Center for Applied Mathematics, Cornell University, USA
**C. Gershenson**
School of Cognitive and Computer Sciences,
University of Sussex, U. K.

*Abstract:*
Search engines perform the task of retrieving information related to the user-supplied query words. This task has two parts; one is finding 'featured words' which describe an article best and the other is finding a match among these words to user-defined search terms. There are two main independent approaches to achieve this task. The first one, using the concepts of semantics, has been implemented partially. For more details see another paper of Marko et al., 2002. The second approach is reported in this paper. It is a theoretical model based on using Neural Network (NN). Instead of using keywords or reading from the first few lines from papers/articles, the present model gives emphasis on extracting 'featured words' from an article. Obviously we propose to exclude prepositions, articles and so on, that is , English words like "of, the, are, so, therefore, " etc. from such a list. A neural model is taken with its nodes pre-assigned energies. Whenever a match is found with featured words and user-defined search words, the node is fired and jumps to a higher energy. This firing continues until the model attains a steady energy level and total energy is now calculated. Clearly, higher match will generate higher energy; so on the basis of total energy, a ranking is done to the article indicating degree of relevance to the user's interest. Another important feature of the proposed model is incorporating a semantic module to refine the search words; like finding association among search words, etc. In this manner, information retrieval can be improved markedly.

## 1. Introduction

Huge collection of data in the form of article, paper, webpages, etc. on various topics are available on the internet. Search engines help us to retrieve information according to our own interest. Algorithms that govern this search operation are getting increasingly complex as complexity in searching terms as well as total volume of available data both are increasing very rapidly. Most evaluation in information retrieval is based on precision and recall using manual relevance judgements by experts. However, especially for large and dynamic document collections, it becomes intractable to get accurate recall estimates, since they require relevance judgements [May, 1999]. We study the relevance point particularly in anther paper by Gershenson at al. (2002).

Search operations are based on extraction of summary from available documents and finding those with good match with user-supplied query or search terms. Lycos, Alta Vista, and similar Web search engines have become essential tools for locating information on the ever-expanding World Wide Web (WWW). Underlying these systems are statistical methods for indexing plain text documents [Mauldin et al., 1994]

Achieving the task of finding degree of relevance of an article according to user's interest has two parts. The first one is to extract features of an article and the other is to match this feature against the user defined search terms. We discuss both the facets in detail along with our understanding of the problem in Section 2. Some existing methods will be discussed therein. Neural or connectionist computation and modeling is an emerging technology with a variety of potential applications such as classification, identification, estimation, etc [Pathak, 1995].
Here we propose a Neural Network (NN) model which will rank an article according to the degree of relevance of it against the user-defined search words. The proposed NN model does both parts of the task associated with such ranking as discussed earlier. Details of the NN model in given Section 3. Refinement of search word in terms of semantics is another important aspect to avoid unrelated retrievals and is discussed in Section 4. A few examples of searching and text summarization will be given at the end of this paper as appendix and are discussed in Section 5. It may be noted that these examples are being reproduced in relation to the paper; any sort of relative comparison is not particularly intended. Finally, we conclude with several pertinent remarks concerning this work

## 2. Extraction of Featured words

Featured words are those words that best describe the paper. Instead of using 'keywords' or extracting first few lines from an article cannot give a good representation of on what the article deals. In fact proper choice of features are most crucial as on these words- the searching is done. Research attempted to this direction carry various names; like 'Text Mining' defined as 'The knowledge discovery process which looks for identifying and analyzing useful information on data which is interesting to users from big amounts of textual data' [Atkinson, 2000] or Information Retrieval which includes tasks like automated text characterization, information extraction, text summarization etc.

Typically, the importance of a word depends both on its frequency in the document under consideration and its frequency in the entire collection of documents. If a word occurs frequently in a particular document, then it is considered salient for that document and is given a high score. In order to select such words, different approaches are in practice. For example, Justin et al. (1997) argues, that as HTML documents are very much 'structured' with tags indicating headings etc. compared to plain text, to weight parameters in the following way. Words in HTML fields should have weights (in the order of maximum to minimum) as follows
i) TITLE
ii) H1, H2, H3 (headlines)
iii) B (bold), I (italics), and BLINK
iv) underlined anchor text

Marcu (1997) represented 7 possible ways to determine the most important parts of a text, for example important sentences in a text i) contain words that are used frequently,(ii) contains words that are used in the title and section headings(iii) are located at the beginning or end of paragraphs iv) are located at positions in a text that are genre dependent—these positions can be determined automatically, through training techniques (v) use bonus words such as "greatest" and "significant" or indicator phrases such as "the main aim of this paper" and "the purpose of this article", while non-important sentences use stigma words such as "hardly" and "impossible" etc.

Fuka et al. (2001) showed that for important-term selection, many different techniques and heuristics that have developed are just a sub-set of more advanced methods originating in the field of pattern recognition.

In this work, we follow the approach of Jennings et al. (1999) to take into account the place of occurrence of a word while considering its weight. For example, a word in the title of an article carries more relevance than one used in the body text. Therefore, different weights are assigned to words according to their place of occurrence. But before that, we have to exclude preposition, article, etc., for example, exclude English words like "of, the, are, so, that, " etc. from the text which are frequently used in any article but are poor candidate to be selected as featured words. The filter shown in Fig. 1 excludes such words. See also the example of summarization, given in appendices, which shows that such exclusion does not affect the 'concept part' of summary of an article.

To form such a featured word list, the article is read first. To read a full paper as input to any processing module would be a heavy load of computation, so a choice of the first few hundred words can serve the purpose -also because beyond this limit, generally technical or scientific notations appear which are not relevant for the present purpose. Determining place of occurrence is a complex task in itself because of different file formats in use; so selection of about first 200 words is sufficient to include the title, author, affiliation and abstract etc. which will be given higher weights than those occurring in the body text.

## 3. Neural Network models

Application of neural network for data compression, feature extraction, and statistical clustering are decade old [Hammerstorm, 1994]. Neural networks are valuable on several counts than traditional programming approach because of the former's learning capacity and its capacity in producing different dynamical states regarding a system which in the present case concerns with total activation energy of excited NN model [Das et al., 2002]. Anderson et al. compared the advantages of using NN over other conventional programmatic approaches in details.

Joachimes (1999) experimentally studied new method leading to conclusive results in a WWW retrieval study and finds which search engine provides better results: Google or MSN (Microsoft Network) Search. Bruza et al. (2000) compared the search effectiveness when using query-based Internet search via Google and Yahoo search engines with the focus to evaluate aspects of the search process.

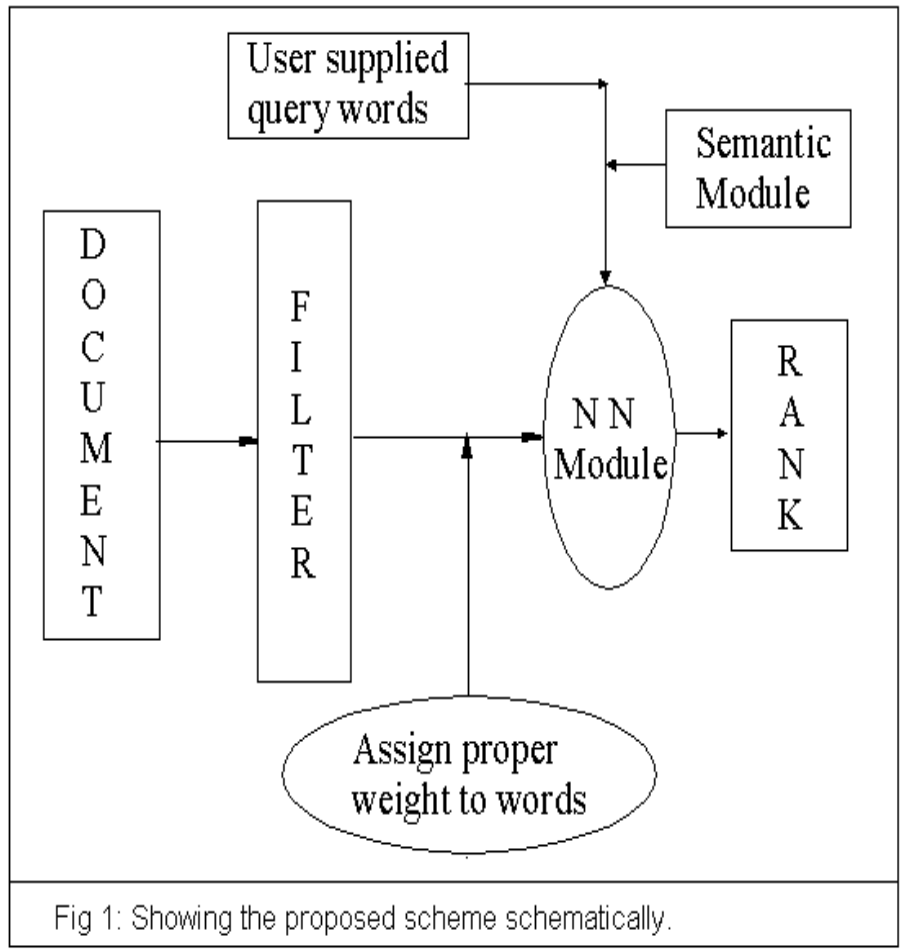

Fig 1: Showing the proposed scheme schematically.

We present here a theoretical neural network learning model with several nodes. The NN model consists of several nodes. Each node is assigned with a word (from the user defined search terms) and pre-assigned equal energy. The model reads an article and if a match between a node and a word is found, that node is fired and gets a higher energy. Here the strength of energy change depends on place of occurrence of the word in the article also. The process of firing will continue until the network settles down to an equilibrium state in accordance to its nodes. So finally we have a set of active nodes and we take the article ranking as the sum of the energy of all the active nodes. An article with higher energy clearly contains more of the search words in its featured word list. So this ranking will indicate the degree of association of the article and user's interest.

So the proposed NN model does both parts of the task associated with ranking an article according to user's interest as discussed earlier.

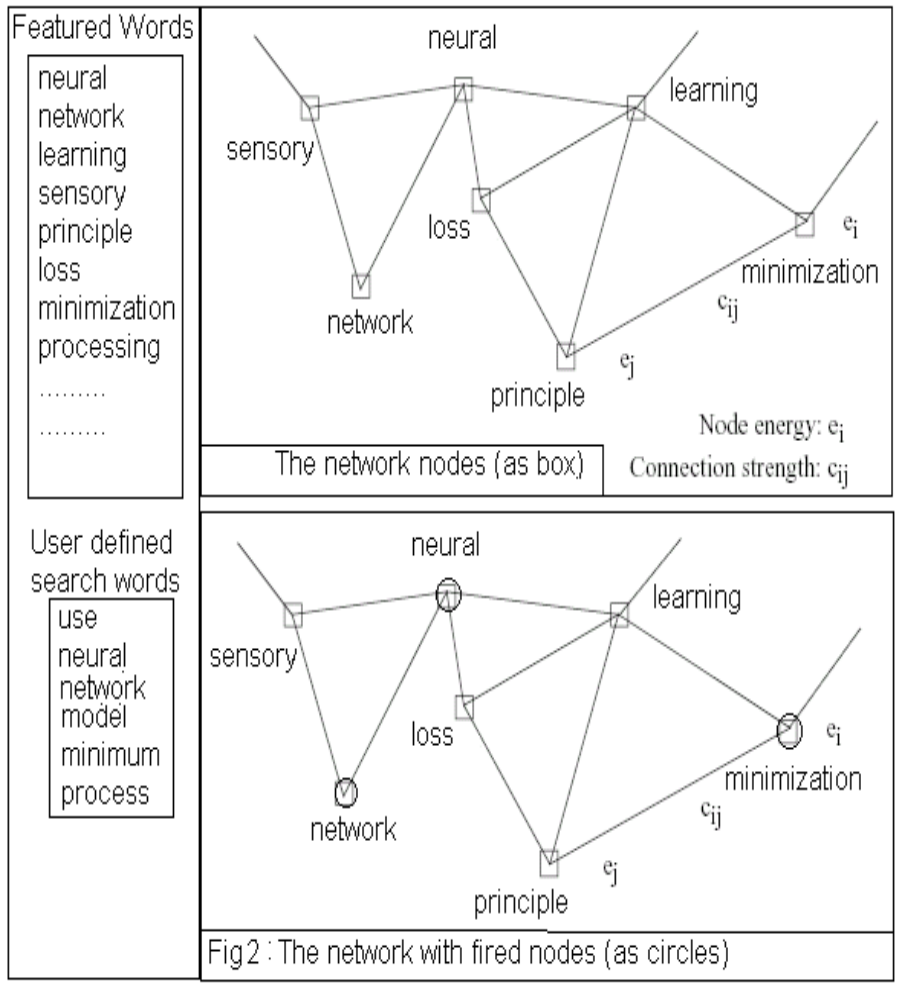

Fig 2 : The network with fired nodes (as circles)

In our network model the strong linking of nodes represents the close relationship of words and their meanings (the issue of refining search words to clarify meaning is described in the section), and this restricts the connectivity of the network. Thus the number of nodes is restricted in that each node represents part of a user interest, and the connectivity is restricted in the sense that a connection is only established among the featured words.

## 3.1 Specific example of Complexity centered web

We like to give a specific example of how the present model can be used for developing a complexity based specialized media. In another paper of our group [Gerhenson et al., 2002] being presented in this conference, concepts and advantages of using specialized media to search WWW for information on specific scientific points (e.g. complexity) have been discussed, in comparison to using general (e.g. Goggle) or semi-specialized (e.g. NEC Research Index) media.

From the database consisting of issues published in last one year of Complexity Digest- a weekly science web-journal focussed on the study of complexity, most commonly occurring seventy words have been found (Gerhenson, ibid.). We can take most commonly used 25 words form this list and substitute them with user defined search words in the proposed model (refer to Fig. 1), keeping in mind that larger number will make the NN model too robust. Now the model has the task of finding match of these 25 words with featured word list of any article as discussed earlier. By this way, we can achieve the task of developing a specialized media for the complex community.

## 4. Refinement of search words

The search engines, at its core of functioning, in order to fetch increasing number of results refine search words. For example, searching for 'physics' includes 'physical chemistry' 'biophysics' 'physical strength', etc. Such refinements can sometimes drift far from the user's interest; sometimes producing meaningless results. In our proposed scheme we have included a semantic module for possible refinement of search words. The purpose is to include words for searching which are not present by the user supplied words. For example, search for word like 'disease' to be refined to include 'disease, illness, ailment' etc. Although it is a completely different domain of research to find the contexts and concept (called the 'meta-data'); some part can be offered off-hand and being presented in the other paper of our group dealing with semantics and ontology creation for WWW sources.

Another challenge is associating words used in such searching, called association". For example, words like 'neural' and 'physiology' can associate to 'neurophysiology' to give more comprehensive results. Existing search implementations largely fail at such situations. Using concept of clusters can solve this problem whereby related words are collected in a group and are activated by any of the members of the group. Agarwal (1995) gave such examples of semantic class generation. Additionally, the concept of clustering in semantic terms can be applied for a more meaningful search and retrieval of relevant results.

## 5. Examples

One can find the limitation of existing algorithms for search operation easily. For example of Google/Yahoo search for the word "ATIN" (name of one of the authors) has returned results with the word "LATIN" also.

In appendix A, we have reproduced an arbitrarily selected article, summarized the text two times independently with Microsoft Office summarizer and Copernic software. Comparison shows that MS summarization stresses on meaningful sentences where Copernic makes a concept list as well as a summary. In both cases, frequently used English words that we propose to filter are retained. This is an unnecessary computational overhead.

Referring to Appendix B where the original text is (manually) filtered and same two summarization tools are applied. Comparing concept list produced by Copernic as given in Appendix A and B, it is clear that excluding such words does not hamper process of concept building. This is an important justification of using filter in our model (refer to Fig. 1). It is also seen from Appendix B that summarization of filtered text results in something meaningless. There lies the need of NN to make a ranking in retrieving information from the article.

## Conclusion:

Extracting relevant features from text is an important challenge. In the present work, we showed that neural models can be used to preprocess an input article and match it with user-defined search terms. Though theoretical, this method can play an important role towards addressing the indicated challenge, as it is expected to lead to marked improvements in the search algorithms employed on the Internet.

**Acknowledgements**: We are grateful to Prof. Gottfried Mayer Kress, Penn. State Univ.,U.S. for giving important suggestions at various stages of this work. C. G. was partly supported by the CONACYT of México and by the School of Cognitive and Computer Sciences of the University of Sussex.

**Appendix A**

--------------(Original Article, First 200 hundred word chosen)-----------

Minimization of Information Loss through Neural Network Learning

M. D. Plumbley

Centre for Neural Networks, Department of Mathematics, King's College London, Strand, London, WC2R 2LS, UK

May 18, 1999

Abstract

Information-transmitting capability of such a neural network is limited both by constraints, such as the number of available units in a particular layer, and by costs, such as the average power used to transmit the information. In this article, we explore the concept of minimization of information loss (MIL) [2] as a target for neural network learning in this context. MIL is closely related to Linsker's Infomax principle [1]. By relating MIL to more familiar supervised and unsupervised learning procedures such as Error Back Propagation (`BackProp') [7] and principal component analysis (PCA) [1], we show how it can be used as a lingua franca for learning in all stages of a neural network sensory system.

1 Introduction
In recent years, a number of authors have used concepts from Information Theory to develop or explain neural network learning algorithms, particu-larly in sensory systems [1, 2, 3, 4, 5]. A neural network in a sensory system
is thought of as part of a communication system, transmitting Shannon
Information [6] about the outside world to further processing stages. The

------ Above summarized by MS Office97   (49 words; Original document 196 words).
Minimization of Information Loss through Neural
Network Learning
Strand, London, WC2R 2LS, UK
Information-transmitting capability of such a neural network is limited both all stages of a neural network sensory system.
Theory to develop or explain neural network learning algorithms, particu- A neural network in a sensory system
-------------same text summarized by Copernic Summarizer-----------
**Concepts:**
> PCA, lingua franca, network learning algorithms, communication system, transmitting Shannon, processing stages, minimization, loss, London, transmit, principle, MIL, sensory system, network learning, neural network.
**Summary:**
> and by costs, such as the average power used to transmit the information loss (MIL) [2] as a target for neural network learning in this context.is closely related to Linsker's Infomax principle [1].
> all stages of a neural network sensory system.
> Information [6] about the outside world to further processing stages.

**Appendix B**
------------Filtered Original Text (see Fig. 1)  --------------------------
Minimization of Information Loss through Neural Network Learning
M. D. Plumbley
Centre for Neural Networks, Department of Mathematics, King's College London,
Strand, London, WC2R 2LS, UK
May 18, 1999

Abstract
Information-transmitting capability neural network limited both constraints number available units particular layer, costs, such average power used transmit information. article explore concept minimization information loss (MIL) [2] target neural network learning context. MIL closely related Linsker's Infomax principle [1] relating MIL  more familiar supervised unsupervised learning procedures Er- ror Back Propagation (`BackProp') [7] principal component analysis (PCA) [1], show how used lingua franca learning  all stages neural network sensory system.

1 Introduction
recent years number authors used concepts Information Theory develop explain neural network learning algorithms, particu- larly in sensory systems [1, 2, 3, 4, 5].

neural network sensory system thought part communication system transmitting Shannon Information [6] about outside world further processing stages

-------Above summarized with MS Office97 (32 words; Original 172 Words)
Network Learning
Information-transmitting capability neural network limited both
article explore concept minimization information
loss (MIL) [2] target neural network learning context. MIL
all stages neural network sensory system.

neural network sensory system
---------Same text summarized by Copernic Summarizer

**Concepts:**

neural network, network learning, sensory system, MIL, principle, transmit, London, loss, minimization, processing stages, system transmitting Shannon, communication system transmitting, network learning algorithms, lingua franca learning, PCA.

**Summary:**

costs, such average power used transmit information.